\documentclass[11pt,letterpaper]{article}

\usepackage[margin=1in]{geometry}
\usepackage[T1]{fontenc}
\usepackage[utf8]{inputenc}

\usepackage{amsmath}
\IfFileExists{newtxtext.sty}{\usepackage{newtxtext,newtxmath}}{}

\usepackage{graphicx}
\usepackage{booktabs}
\usepackage{multirow}
\usepackage{array,makecell}
\usepackage{longtable}
\usepackage{adjustbox}
\usepackage{tabularray}   \usepackage{siunitx}
\usepackage{xcolor}      \usepackage{caption}
\usepackage{subcaption}  \usepackage{authblk}      
\usepackage[american]{babel}
\usepackage{csquotes}
\usepackage[style=apa,backend=biber]{biblatex}
\DeclareLanguageMapping{american}{american-apa}

\usepackage[hidelinks]{hyperref}

\AtBeginDocument{\let\cite\parencite}

\makeatletter

\newcommand{\articletype}[1]{}
\newcommand{\authormark}[1]{}
\newcommand{\received}[1]{}
\newcommand{\revised}[1]{}
\newcommand{\accepted}[1]{}
\newcommand{\journalname}[1]{}
\newcommand{\journalvolume}[1]{}
\newcommand{\journalnumber}[1]{}
\providecommand{\doi}[1]{}

\newcommand{\email}[1]{\texttt{#1}}

\newcommand{\address}[2][]{\affil[#1]{#2}}

\newcommand{\@corrtext}{}
\newcommand{\corres}[1]{\gdef\@corrtext{#1}}

\newcommand{\@abstracttext}{}
\renewcommand{\abstract}[2][Abstract]{\gdef\@abstracttext{#2}}
\newcommand{\@keywordstext}{}
\newcommand{\keywords}[1]{\gdef\@keywordstext{#1}}

\newcommand{\printfrontmatter}{  \begingroup
    \ifx\@corrtext\@empty\else
      \renewcommand{\thefootnote}{}      \footnotetext{\@corrtext}      \renewcommand{\thefootnote}{\arabic{footnote}}    \fi
    \begin{center}    \begin{minipage}{0.92\textwidth}      \small
      \noindent\textbf{Abstract.}\ \@abstracttext\par
      \vspace{0.8em}
      \noindent\textbf{Keywords:}\ \@keywordstext\par
    \end{minipage}    \end{center}    \vspace{1em}
  \endgroup}

\makeatother

\date{}

\title{Development of Vision-Language Model–based GNSS Spoofing Detection for Autonomous Vehicle Navigation}

\author[1]{Mohammed Aldeen*}
\author[2]{Muhammad Sami Irfan}
\author[2]{Sagar Dasgupta}
\author[1]{Long Cheng}
\author[2]{Mizanur Rahman}
\author[3]{Mashrur Chowdhury}

\authormark{ALDEEN \textsc{et al.}}

\address[1]{School of Computing, Clemson University, Clemson, SC, USA}
\address[2]{Department of Civil, Construction \& Environmental Engineering, The University of Alabama, Tuscaloosa, AL, USA}
\address[3]{Glenn Department of Civil Engineering, Clemson University, Clemson, SC, USA}

\corres{*Mohammed Aldeen, School of Computing, Clemson University, Clemson, SC, USA.\\ \email{mshujaa@clemson.edu}}

\begin{document}

\abstract[Abstract]{Autonomous vehicles (AVs) depend on Global Navigation Satellite Systems (GNSS) for localization and navigation, making them vulnerable to spoofing attacks that can covertly redirect vehicles or induce unsafe maneuvers. In this paper, we develop the first Vision-Language Model (VLM)-based framework for GNSS spoofing detection for autonomous vehicles by fusing front-camera visual data with in-vehicle sensor readings (e.g., speed, acceleration, yaw rate) against GNSS-derived maneuvers. Our approach introduces a three-stage fine-tuning process that first grounds visual cues, and then calibrates sensor data within a shared semantic space to detect discrepancies between predicted and GNSS-derived maneuvers across three attack scenarios. We also generated an independent real-world dataset by driving an instrumented vehicle on public roads in Tuscaloosa, Alabama, equipped with time-synchronized GNSS, IMU, and camera logs to validate cross-regional generalization of our fine-tuned model on unseen data from training data. On this dataset, we then generated intelligent spoofing attacks, including trajectory mirroring with road-network snapping for wrong-turn attacks, position freezing for overshoot scenarios, and drift generation for stop attacks. On this validation dataset, the zero-shot VLMs baseline F1-score ranges from 23\% to 32\%, whereas our fine-tuned model achieves an F1‑score ranging from 94\% to 95\%. Results show that our VLM-based approach correctly classified every wrong‑turn and stop attacks, and attains 88\%–93\% accuracy for overshoot attacks. Furthermore, we introduce an adaptive inference policy that reduces VLM invocations to 14\% ($ \sim $ 86\% computational reduction) and yields 65ms–73ms per 4s window. These results point to a practical, on-road layer of defense that complements signal-level integrity checks with the use of VLMs.}


\maketitle
\printfrontmatter

\noindent\textit{Preprint. This manuscript has been submitted to
NAVIGATION, Journal of the Institute of Navigation
(\url{https://navi.ion.org}) and is currently under peer review.}
\vspace{1.5em}

\section{Introduction}
The rapid advancement of autonomous vehicle (AV) technology has fundamentally transformed modern transportation systems, with Global Navigation Satellite System (GNSS) serving as a critical component for precise localization and navigation \cite{chen2023gnss}. As autonomous vehicles increasingly rely on GNSS data for path planning \cite{zhao2019detection}, route optimization \cite{nowak2017dynamic}, and safety-critical decision-making \cite{filip2024continuity}, the integrity and authenticity of GNSS signals have become paramount concerns for vehicle safety and operational reliability \cite{filip2024continuity}. Historically, civilian GNSS signals lacked built-in authentication. While recent schemes such as the operational Galileo OSNMA and the GPS CHIMERA protocol introduce cryptographic protections at the signal level~\cite{simon2024galileo,anderson2017chips}, the vast majority of deployed autonomous vehicles rely on legacy hardware incapable of processing these new authentication protocols, and therefore cannot benefit from them~\cite{taoglas2025osnma,gsc2026osnma}. Consequently, autonomous navigation systems remain vulnerable to sophisticated attacks where adversaries broadcast false satellite signals to deceive receivers~\cite{wu2020spoofing,psiaki2016gnss}. Unlike jamming attacks that deny service, spoofing attacks operate covertly, manipulating an autonomous vehicle's position and trajectory without triggering conventional detection mechanisms~\cite{shen2020drift}. The proliferation of low-cost software-defined radios has further lowered the barrier for conducting such attacks \cite{zeng2018all}, enabling attackers to force deliberate route deviations and guide vehicles to unintended destinations while maintaining consistency with surrounding road geometry.

Current GNSS spoofing detection approaches exhibit three fundamental limitations that leave autonomous vehicles vulnerable to sophisticated attacks.  First, traditional receiver-based integrity checks that operate on processed GNSS measurements, including Receiver Autonomous Integrity Monitoring (RAIM) and Doppler-consistency checks, can be bypassed by slow-drift or “carry-off” attacks that remain synchronized with legitimate signals~\cite{kujur2024optimal}. Second, modern multi-sensor fusion systems that combine GNSS with an Inertial Measurement Unit (IMU), odometry, and LiDAR data also remain vulnerable to  attacks that simultaneously manipulate GNSS outputs in a manner that remains consistent with other onboard sensor readings, preventing fusion-internal alarms from triggering. For instance, the FusionRipper attack demonstrated successful exploitation of timing vulnerabilities in production autonomous vehicle algorithms without triggering internal consistency checks~\cite{shen2020drift}.  Third, recent approaches validate GNSS by comparing it against independent visual or geometric estimates, such as visual-inertial odometry (VIO), visual SLAM, or camera-to-map alignment. These methods provide important cross-domain consistency checks, but they primarily operate in a metric or geometric space. They compare GNSS-derived motion against quantities such as pose, displacement, heading, trajectory, visual feature correspondence, or map alignment.  As a result, they depend on highly accurate sensor calibration and precise temporal alignment across modalities. VIO and visual SLAM are mature methods that build local landmark maps online within a SLAM framework capable of real-time operation,  including ORB-SLAM3 and VINS-Mono/VINS-Fusion~\cite{qin2018vins, campos2021orb}, yet they are prone to cumulative drift over long trajectories and degrade under adverse visual conditions such as low texture, low light, glare, rain, fog, motion blur, or shadows~\cite{tabassum2025comparative}. On the other hand, camera-to-map alignment avoids tracking drift but relies heavily on expensive prior maps that require frequent updates. Furthermore, its visual matching performance degrades significantly when lighting, weather, or seasonal changes alter the scene's appearance ~\cite{elghazaly2023high, jiang2024seek}. Section~\ref{sec:related} provides a more detailed review of GNSS spoofing attacks and existing spoofing-detection approaches.

The emergence of Vision Language Models (VLMs) presents a promising opportunity to address these limitations from a different perspective. Instead of validating GNSS through precise geometric correspondence or direct camera-to-map feature matching, VLMs are able to perform validation in a semantic behavior space. This behavior-space validation is central to navigation-level spoofing: whether the semantic behavior implied by GNSS is consistent with the behavior independently observed by the vehicle. A spoofed GNSS trajectory can remain mathematically smooth, road-consistent, and geometrically plausible while still implying an unsafe or incorrect driving behavior, for example, implying a left turn while the camera and vehicle dynamics indicate a right turn, or indicating a stationary state while the vehicle continues to move. These sophisticated attacks require validation at the behavior level, not only at the metric trajectory level. Therefore, VLMs are suited to this semantic validation because their cross-modal semantic grounding enables them to map visual scene evidence and numeric sensor changes into a shared representation space, producing behavioral descriptions that can be directly compared against GNSS. This semantic comparison is generally less brittle than VIO or camera-to-map approaches because it does not depend on exact pixel-to-pixel features, where changes in lighting, weather, or season can weaken direct visual matching. This distinction does not mean that the VLM is unaffected by adverse visual conditions. Rather, it is less dependent on dense visual correspondence because the behavior estimate is also supported by vehicle-side telemetry, including speed, longitudinal acceleration, and yaw rate. Therefore, in this work we propose a VLM method to  evaluate GNSS at the semantic behavior level rather than the metric localization level. In addition, standard image classifiers and sensor-based models each operate on a single modality and cannot produce this cross-modal semantic output~\cite{dasgupta2022sensor, jiang2024seek}.

This semantic validation paradigm is directly motivated by a rapid industry-wide shift toward multimodal Vision-Language Models (VLMs) and Vision-Language-Action (VLA) architectures for next-generation autonomous driving. While VLMs provide the foundational reasoning by processing camera and sensor streams, VLA models actively translate this visual understanding into vehicle control and physical driving actions. The commercial space, for example, Waymo has developed and released EMMA (End-to-End Multimodal Model), built directly on multimodal LLM foundations (Gemini-class) to map raw camera data to planning, perception, and road understanding outputs~\cite{hwang2024emma}. Also, NVIDIA launched the Alpamayo family of open reasoning VLA models designed for robotaxis autonomy~\cite{wang2025alpamayo}. Academic research has undergone the exact same shift; systems such as  OpenDriveVLA~\cite{zhou2026opendrivevla}, AutoVLA~\cite{zhou2026autovla}, and Reasoning-VLA~\cite{zhang2025reasoning}, demonstrate that multimodal foundation models are becoming a dominant architecture for next-generation autonomous driving. This trend strengthens the practical relevance of a VLM-based GNSS spoofing detector because the detector evaluates GNSS at the same decision-relevant behavior level at which future AV autonomy stacks increasingly operate.

At the same time, applying VLMs to real-time GNSS spoofing detection introduces important technical challenges. Current large VLMs typically contain billion-scale parameters, making inference computationally intensive and creating latency bottlenecks for safety-critical applications that require sub-second response times~\cite{aldeeniv}. Moreover, effective spoofing detection still requires synchronized camera, IMU, and GNSS inputs so that all modalities describe the same driving moment. Even when such alignment is available, existing VLMs lack robust temporal reasoning capabilities between the video frames that could reduce reliability on time-sensitive spoofing events~\cite{cai2024temporalbench}. Also, VLMs exhibit reliability issues due to hallucinations and  overconfident predictions, especially when encountering driving scenarios that differ from the training data.

To address these challenges, this paper presents the first VLM-based framework for GNSS spoofing detection in autonomous vehicles. Our framework identifies discrepancies between actual vehicle maneuvers and GNSS-reported navigation data by combining camera visual cues with IMU data. In order to overcome the computational bottleneck of VLM, we develop an adaptive inference optimization that reduces VLM invocations through sensor-based gating that triggers VLM inference only when sensor readings indicate maneuver changes. Our approach also introduce a three-stage fine-tuning process that explicitly aligns visual and sensor data within a shared semantic space, ensuring multimodal understanding across time-critical events. To mitigate reliability concerns under unfamiliar conditions, we generate an independent real-world evaluation dataset in Tuscaloosa, Alabama with synchronized GNSS, IMU, and camera data under both attack-free and simulated spoofing scenarios. Our detection framework was evaluated across three critical attack scenarios: wrong-turn attacks that manipulate turning directions, overshoot attacks that freeze the reported vehicle position during movement, and stop attacks that represent vehicle movement while stationary.

\textbf{Contributions:} The primary contributions of this study are threefold:
\begin{itemize}
    \item We develop the first VLM-based framework for GNSS spoofing detection in autonomous vehicles by fusing front-camera visual data with in-vehicle sensor readings (speed, acceleration, yaw rate).  Our approach introduces a three-stage fine-tuning process that aligns visual and sensor data within a shared semantic space to detect discrepancies between predicted and GNSS-derived maneuvers across three attack scenarios.

    \item We generate an independent real-world validation dataset by driving an instrumented vehicle on public roads in Tuscaloosa, Alabama, equipped with time-synchronized GNSS, IMU, and camera logs to assess cross-regional generalization on \textbf{unseen} data from training data. We also generate road attacks, including trajectory mirroring with road-network snapping for wrong-turn attacks, position freezing for overshoot scenarios (\textit{i.e.,} the spoofed GNSS coordinate is held constant while the vehicle continues to move), and synthetic drift generation for stop attacks (\textit{i.e.,} a slow synthetic latitude/longitude drift is injected while the vehicle is actually stationary).

    \item We introduce an adaptive inference policy that significantly reduces VLM computational overhead while maintaining detection accuracy. This policy monitors speed, yaw rate, and longitudinal acceleration to schedule VLM inference only when maneuver changes are indicated avoiding continuous model calls.

\end{itemize}

\section{Related Work and Background}\label{sec:related}

The vulnerability of GNSS to spoofing has been recognized for more than two decades, beginning with early assessments by the Volpe National Transportation Systems Center~\cite{infrastructure2001vulnerability}. Autonomous vehicles, which depend heavily on GNSS for precise positioning and navigation, are susceptible to intentional threats such as spoofing. Spoofing attacks are a more sophisticated form of attack in which an adversary generates counterfeit GNSS signals, deceiving the receiver into computing false positions or routes \cite{shen2020drift}. For a successful spoofing attack, an attacker often requires knowledge of the target vehicle’s current location, intended route, and destination. For example, researchers have demonstrated that even subtle injections of spoofed signals can mislead a vehicle, causing it to follow an incorrect route \cite{zeng2018all}. Some spoofing techniques are specifically designed to target integrated navigation systems \cite{geng2022research}. In this study, we focus on sophisticated GNSS spoofing attacks targeting autonomous vehicles.

Prior defenses fall into three broad categories: receiver-level integrity checks, RF-level signal defenses, and software-based cross-sensor consistency checks~\cite{meng2022survey}. Receiver-level methods, such as RAIM, evaluate the consistency of processed satellite pseudorange measurements. These methods are useful for identifying inconsistent satellite observations, but a sophisticated spoofer can generate mathematically self-consistent false measurements that pass residual-based checks~\cite{workgroup2024gps}. RF-level defenses operate closer to the received signal and monitor indicators such as received power, carrier-to-noise density ratio, correlation peak distortion, or angle of arrival. For instance, single-antenna receivers can monitor signal power and carrier-to-noise ratios to detect the abnormal power surges introduced when a spoofer overpowers legitimate satellites \cite{lo2021developing}. However, relying exclusively on single-antenna checks is problematic in dense urban environments, where signal multipath causes extreme power fluctuations and high false alarm rates, and advanced coherent spoofing can bypass them entirely. To detect these coherent attacks, multi-antenna spatial defenses can improve robustness by comparing the physical arrival angles of the received signals~\cite{konovaltsev2013autonomous}. However, they require synchronized phase measurements and specialized antenna hardware that are not always available on deployed autonomous vehicle platforms~\cite{mao2023gnss}. 

For that reason, much of the recent literature has shifted toward software-deployable cross-domain validation methods. For example, Kujur et al. \cite{kujur2024optimal} present an Inertial Navigation System (INS) monitor that accumulates small discrepancies in an integrated GNSS/INS solution to detect the subtle tracking errors induced by spoofing even centimeter-level slow deviations less than $60$ seconds. Recent research expands on this by fusing GNSS with IMUs, odometry, and LiDAR to actively forecast the vehicle's next physical position and compare it against the GNSS reading \cite{dasgupta2020prediction}. These methods improve robustness, but they still validate primarily in a numerical or geometric space. They ask whether GNSS is consistent with another estimate of position, displacement, trajectory, heading, or map correspondence. As a result, a sophisticated spoofer can evade these metric fusion filters entirely by keeping the fake GNSS trajectory consistent with the vehicle’s actual motions and the physical road network. For example, the spoofed trajectory may imply an incorrect turn direction, a frozen position during motion, or false motion while the vehicle is stationary \cite{zeng2018all, song2021credible}. For instance, recent attacks on multi-sensor fusion localization, such as FusionRipper \cite{shen2020drift} and Ghost Navigator \cite{zhang2025ghost}, demonstrate that adversaries can exploit timing and filter vulnerabilities to manipulate fused outputs, which shows that internal numerical consistency alone may not be sufficient for robust GNSS integrity monitoring. These studies motivate the need to evaluate GNSS not only at the metric trajectory level, but also at the behavior level: whether the maneuver or motion state implied by GNSS agrees with what the vehicle independently observes through camera frames and vehicle-side sensors.

Since the remaining weakness in prior defenses is semantic behavior rather than purely geometric/numerical, the next relevant line of work is on models that can infer driving behavior across modalities. Recent autonomous driving literature has explored semantic and cognitive frameworks using Large Language Models (LLMs) and multi-modal Vision-Language Models (VLMs) \cite{sriram2019talk,omama2023alt}. A VLM is a deep learning architecture designed to bring together different types of data, such as images and text, into a unified semantic space where they can be understood together. As structurally illustrated in Figure~\ref{fig:overview}, it typically combines three key components: a deep vision encoder that extracts rich spatial features from raw camera frames, a large language model at its core that handles contextual reasoning, and a cross-modal projection layer that converts visual representations into textual tokens the language model can work with directly \cite{zhang2024vision, xia2024new}. Because text-only LLMs cannot directly see or understand the real driving environment through cameras and sensors, multi-modal VLMs have emerged to resolve this limitation by natively coupling sensory telemetry and sequential imagery~\cite{zhang2024vision}. Recent advances in VLMs, such as CLIP~\cite{DBLP:conf/icml/0001LXH22}, BLIP~\cite{DBLP:conf/icml/0001LXH22}, and LLaVA~\cite{DBLP:journals/corr/abs-2304-08485}, demonstrate that neural networks jointly trained on images and text can achieve sophisticated scene understanding, semantic grounding, and cross-modal reasoning capabilities. These architectures can interpret complex visual scenes, identify landmarks, read traffic signs, and understand spatial relationships in real-time~\cite{jiang2024senna}. Their contextual understanding leverages textual information to guide visual scene analysis~\cite{zhao2024enhancing} using multi-modal feature extraction to fuse features from both modalities \cite{bain2021frozen}. VLMs have also shown significant promise for AVs across perception, control, and motion planning, giving vehicles a more complete environmental understanding~\cite{cui2024receive} and strengthening AV cybersecurity defenses \cite{aldeeniv}.

Taken together, prior spoofing defenses verify whether GNSS is geometrically consistent with other sensors. A complementary layer of defense is needed to audit the semantic behavior implied by GNSS against independent camera and in-vehicle sensor evidence. In this work, we shift the validation question from metric consistency to semantic behavior consistency. Rather than asking only whether GNSS is numerically consistent with vehicle-side estimates of position, displacement, or trajectory, we ask whether the maneuver or motion state implied by GNSS is semantically consistent with what the vehicle observes through synchronized camera frames and in-vehicle sensors. VLM reasoning is useful for this behavior-level check because it can convert the camera/sensor stream into a semantic description of the current driving episode, such as \textit{``the vehicle is turning right,''} \textit{``the vehicle is stopping,''} or \textit{``the vehicle is continuing straight.''} The detector then compares this camera/sensor-derived behavior with the GNSS-derived maneuver or motion state. When these behavioral descriptions disagree, the system flags a spoofing event.

On the other hand, VLMs face critical limitations, including computational expense hindering real-time deployment~\cite{zhang2024minidrive}, inadequate spatial reasoning and numerical precision for trajectory predictions~\cite{jiang2024senna}, and reliability issues from hallucinations in safety-critical scenarios~\cite{xing2024autotrust,wang2024rac3}. Because of these challenges, a VLM cannot be deployed as a direct plug-and-play replacement for primary metric localization; rather, it must be carefully tuned. To define the scope of our study, we target spoofing cases in which the GNSS trace changes the semantic driving behavior. Localization-drift attacks, or attacks that preserve the same high-level semantic behavior across GNSS, camera, and in-vehicle sensors, remain outside the primary scope of this framework. Also, attacks that compromise camera data, in-vehicle sensors, or all fused perception outputs so that GNSS and onboard observations remain semantically consistent are outside the primary scope of this work. In this work, the choice of a VLM-based approach does not imply that signal-level methods or LiDAR-based localization are unnecessary. Rather, the proposed method addresses a different detection target. Signal-level methods test the authenticity of the received GNSS signal, and LiDAR or VIO-based approaches support geometric localization consistency. The proposed VLM framework tests whether the behavior implied by GNSS is consistent with independently observed vehicle behavior. This behavior-level question is central to navigation-level spoofing because the safety consequence of the attack is often an incorrect maneuver or motion-state decision, not only a coordinate error. We use camera frames together with speed, longitudinal acceleration, and yaw rate because these inputs directly support semantic maneuver and motion-state estimation while avoiding the additional representation complexity of converting raw LiDAR point clouds into VLM-compatible inputs. Incorporating LiDAR into the behavior-estimation module remains a valuable extension for future work.

\section{Datasets} \label{sec:data}

\subsection{CoVLA Dataset} 
Comprehensive Vision-Language-Action Dataset for Autonomous Driving (CoVLA) is a real-world driving corpus collected in Tokyo, Japan that provides time-synchronized forward-facing video, Global Navigation Satellite System (GNSS) readings, six-axis inertial measurements (three-axis linear acceleration and three-axis angular velocity), and Controller Area Network (CAN) vehicle signals \cite{arai2025covla}. All modalities are recorded on a shared clock, enabling frame-level pairing of visual context with vehicle motion and state. The data include instances of common maneuvers (straight, stop, left turn, right turn) along with their corresponding clips and sensor traces. In this work, we used CoVLA dataset solely for training our vision–language model (VLM) so it learns to associate changes across consecutive images with concurrent changes in the sensor traces, rather than model to learn from on other visual elements such as road marking geometry or camera viewpoint.

\begin{figure}[b]
\vspace{-10pt}
    \centering
        \begin{subfigure}[t]{0.53\textwidth}
        \centering
        \includegraphics[width=\linewidth]{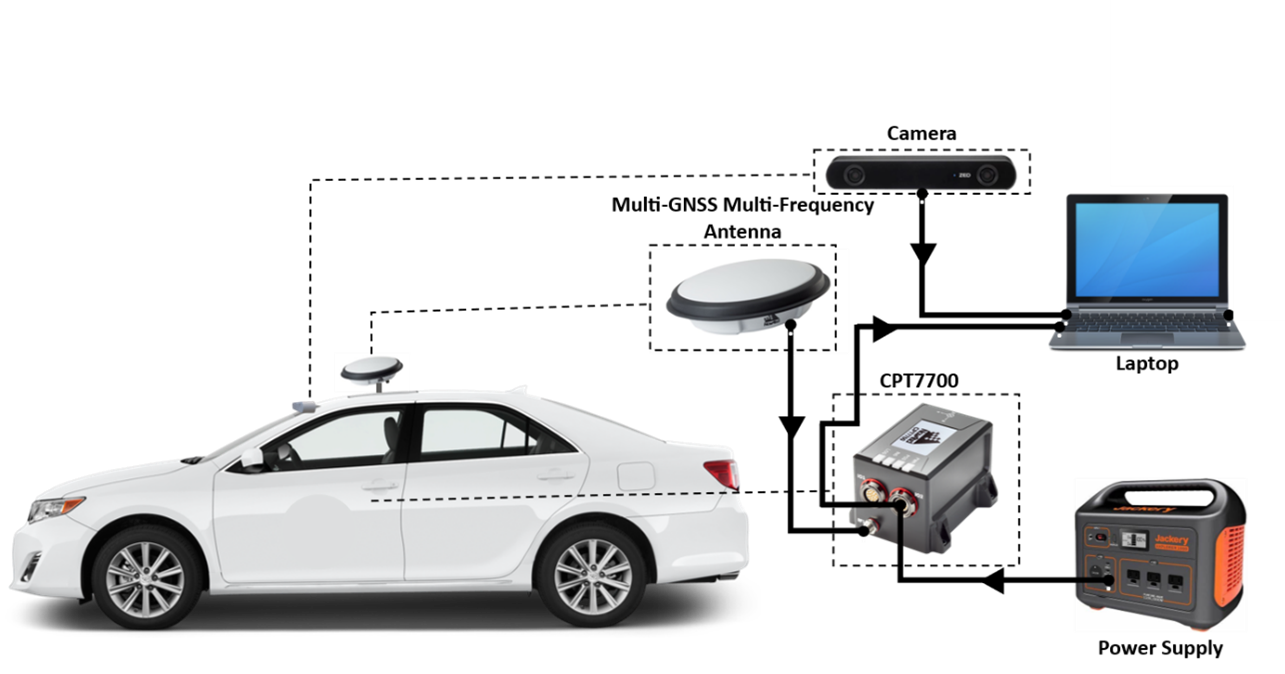}
        \caption{\centering Sensor setup for data collection}
        \label{fig:setup}
    \end{subfigure}
    \hspace{0.02\textwidth}         \begin{subfigure}[t]{0.39\textwidth}
        \centering
        \includegraphics[width=\linewidth]{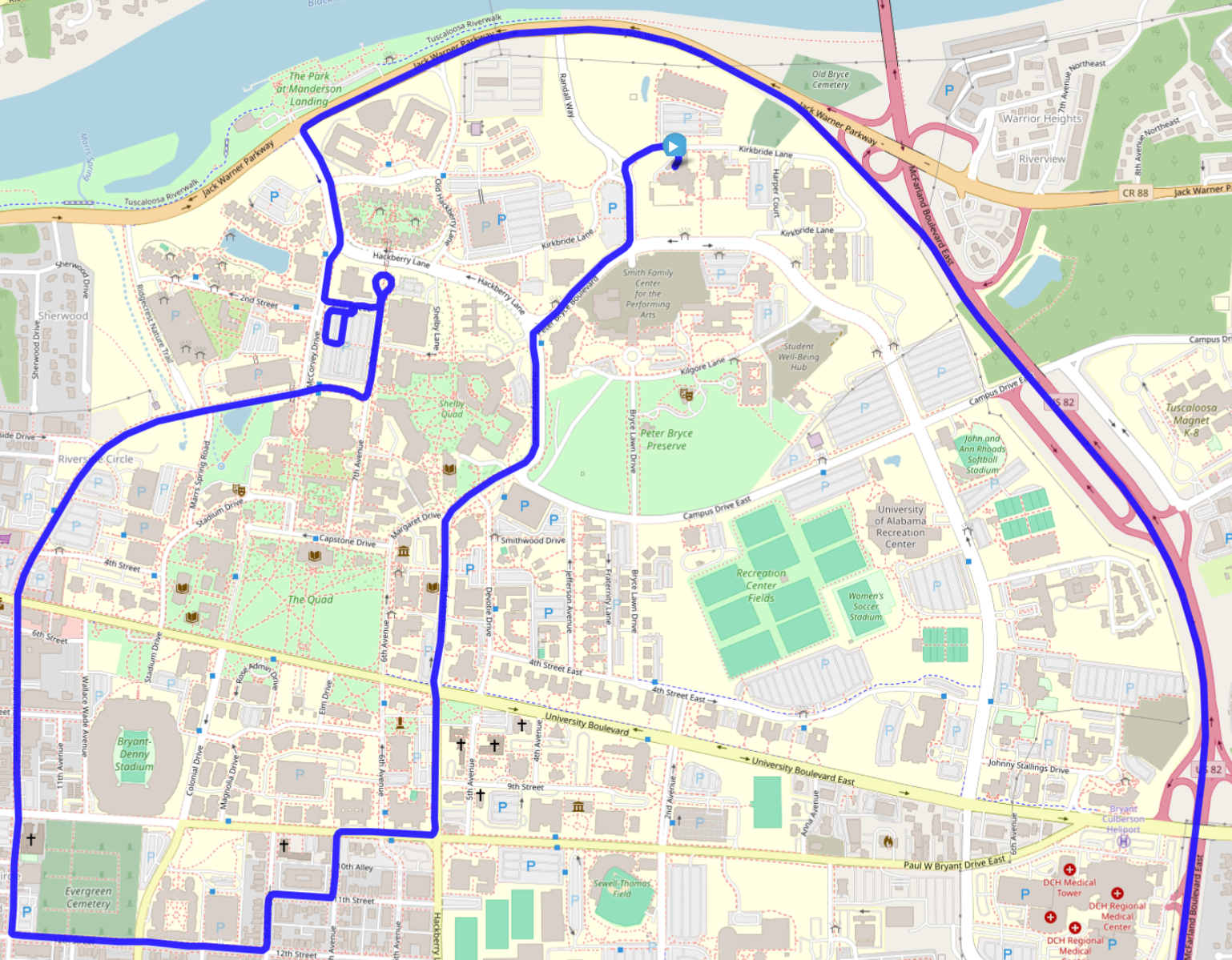}
        \caption{\centering Collected GNSS traces from our data}
        \label{fig:traces}
    \end{subfigure}
   
    \caption{Overview of our data collection process: (a) sensor suite installed on the vehicle and (b) GNSS traces collected during driving.}
    \label{fig:data_collection}
\vspace{-10pt}
\end{figure}
\subsection{Field Collected Dataset} 

As mentioned above, the dataset used for training the VLM is CoVLA, which was collected on roads in Tokyo, Japan. This environment differs substantially from Tuscaloosa, Alabama, USA: vehicles drive on the left side of the road in Japan, lane widths and intersection geometry differ, and signage and pavement markings follow different standards. To evaluate cross-regional generalization, we collected an independent on-road validation dataset in Tuscaloosa using the same sensor suite and data processing pipeline, as our training data from Tokyo. This Tuscaloosa validation dataset was used only for evaluation and was not used to train or tune any component of the proposed method. The purpose of this validation design is to test whether the model learns domain-invariant driving-behavior cues rather than memorizing region-specific visual appearance. The mechanism supporting this transfer is described in Section~\ref{sec:fine_tuning}.

As illustrated in Figure \ref{fig:setup}, the vehicle we used in this study was equipped with a NovAtel CPT7700 GNSS+IMU system (providing 2cm accuracy via TerraStar-L corrections), CAN-bus vehicle speed logging, and a front-facing Stereolabs ZED 2 camera capturing 720p video at 15fps. All sensor data streams, including GNSS (“GPGGA”), velocity (“BESTGNSSVEL”), and IMU (“RAWIMU”) logs, were recorded at 10Hz and timestamped by an onboard computer for precise synchronization. Video frames were extracted and aligned with sensor data using the camera SDK and converted to UTC to ensure temporal correspondence between modalities. Furthermore, this validation data was captured using a different camera hardware than the one used in the CoVLA training dataset. This hardware difference allows us to test whether the features our model learned are tied to one specific camera or can generalize well across moderate changes in sensor hardware. Data was collected while driving the instrumented vehicle around Tuscaloosa on public roads, generating the comprehensive GNSS trajectory traces shown in Figure \ref{fig:traces} and a sample of the collected data with its time synchronization is provided in Table~\ref{tab:honda-sample}.

\begin{table}[h]
\centering
\footnotesize
\caption{Sample data from our collected dataset}
\label{tab:honda-sample}
\begin{tblr}{
  colspec={X[1.1,l]|X[1.3,c]X[1.5,c]|X[c]X[c]X[c]X[1.1,c]X[1.2,c]X[1.2,c]X[1.2,c]|X[0.9,c]},
  row{even} = {white},
  row{odd}  = {bg=black!10},
  row{1,2}  = {bg=black!20,font=\bfseries},
  hline{1,3,Z} = {1pt,solid,black!60},
  hline{2} = {solid,black!40},
  vline{2,4,11} = {1pt,solid,black!60},
  colsep=2.5pt, rowsep=1.5pt
}
\SetCell[r=2]{c} UTC Time
  & \SetCell[c=2]{c} GPS &
  & \SetCell[c=7]{c} In-vehicle Sensors & & & & & &
  & \SetCell[r=2]{c} Frame \\
& Lat.\ ($^\circ$) & Lon.\ ($^\circ$)
& $v$ & $a_x$ & $a_y$ & $a_z$ & $\omega_x$ & $\omega_y$ & $\omega_z$ & \\
190032.5 & 33.219540 & $-$87.538482 & 8.05 & 2.30 & $-$0.85 & $-$9.41 & 0.0203 & $-$0.0016 & $-$0.0448 & 1064 \\
190033.0 & 33.219503 & $-$87.538468 & 8.54 & 1.87 & $-$0.52 & $-$10.28 & 0.0498 & 0.0655 & $-$0.0107 & 1072 \\
190033.5 & 33.219466 & $-$87.538453 & 9.03 & 1.58 & $-$0.06 & $-$10.03 & $-$0.0011 & 0.0183 & 0.0291 & 1079 \\
190034.0 & 33.219426 & $-$87.538438 & 9.47 & 2.19 & 0.40 & $-$9.75 & $-$0.0036 & 0.0168 & 0.0716 & 1087 \\
\end{tblr}
\par\vspace{3pt}
{\footnotesize\raggedright \textit{Note.} $v$ = speed (m/s); $a$ = acceleration (m/s$^2$);
$\omega$ = angular rate (rad/s).\par}
\end{table}

\vspace{-20pt}

\subsection{Data Preparation} \label{sec:data}
\subsubsection{Filtering Driving Clips} 
\begin{figure}[h]
\vspace{-10pt}
    \centering
    \includegraphics[scale=0.74]{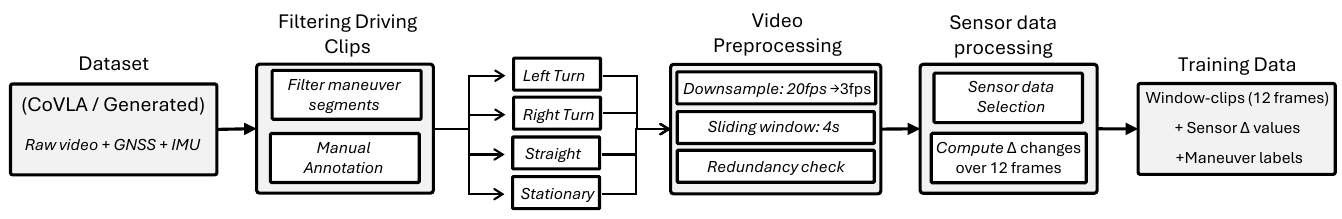}
    \caption{Data preprocessing workflow}
    \label{fig:data_preprocessing}

\end{figure}

Figure~\ref{fig:data_preprocessing} illustrates the complete data preparation pipeline from raw dataset collection through video and sensor preprocessing to generate training-ready window-clips with synchronized multimodal inputs. This preparation pipeline is applied identically to both datasets training dataset (CoVLA) and Tuscaloosa validation dataset. For CoVLA, We first narrowed the clips to only include those clips with driving maneuvers to improve model understanding (i.e., straight driving is visually simpler and less informative). To do that, we used the data  filtered the dataset to include only segments labeled with maneuvers such as \textit{"turning"} and\textit{ "curving"} based on the captions from CoVLA corpus. From this filtered pool, we randomly sampled 100 video clips (each $\sim 30$  seconds at 20 FPS), extracted all frames, and manually annotated each one by assigning the ego vehicle’s maneuver to exactly one of four categories:\textit{ “straight,” }\textit{“stopped,” “turning left,”} or \textit{“turning right"}. Similarly, Tuscaloosa validation dataset follow the same procedure, where frames were extracted and each frame was manually labeled using the same four maneuver categories we used for CoVLA.

\subsubsection{Video Processing}\label{sec:video}

Our goal is to infer maneuvers from the captured videos, we need first separate actual turns from the natural curvature of straight road segments. Road geometry can be complex, even so-called “straight” roads often include curves that require subtle yaw adjustments. One straightforward approach is to feed raw camera frames directly into the model to understand if the car is making a turn; however, it may interpret these road curves as intentional turns given from the camera angle, when in reality, the vehicle is simply following the road’s shape. 

seTo avoid this pitfall, we first segment the continuous recording into short clips, each clip corresponding to a single driving action: \textit{vehicle is going straight, making a left turn, making a right turn, or stopping}—so that the label reflects the driver’s intent rather than the asphalt’s shape. However, feeding these videos directly into the VLM could \textit{(i)} inflate memory and computation \textit{(ii)} overwhelm the model with near-identical frames, making it harder for the model to understand the action by the driver. Thus, within each clip we down-sample the original $f_{\text{in}}=20\;\text{fps}$ stream to
$f_{\text{out}}=3\;\text{fps}$ and cap the temporal time span at $T=4\;\text{s}$,
yielding exactly twelve bearing frames, while preserving uniform temporal coverage. After down‑sampling the video to 3 FPS, we construct a \emph{window‑clip} at every frame by packaging the previous 12 frames (4s) into one sample (stride = 1 frame); in other words, each new frame yields a new window‑clip consisting of the preceding 12 frames. The window‑clip inherits the anchor frame’s maneuver label. This rolling‑window expansion produces approximately 1,000 overlapping window‑clips from the original 100 raw clips, which we use for all model training and testing. To avoid confusion, throughout the paper we use \emph{raw clip} to mean the original 30‑s sequence and \emph{window‑clip} to mean the 4‑s sample created by the sliding window.

We then apply a
light-weight redundancy check: inspired by the two-stage keyframe procedure in \cite{liang2024keyvideollm}, we feed the
twelve candidate images through a frozen CNN, and compute the cosine similarity between the frames. If a frame has a cosine similarity score greater than 0.9 with any other frame in the same scene, it is considered a near-duplicate and is removed. We explicitly sample and process individual frames rather than feeding raw short video into the VLM because video inputs may exceed the model’s fixed context window or internal frame budget, causing it to skip or drop important maneuver frames automatically \cite{liu2025keyframe}.  Also, given that VLMs have limited temporal reasoning over long or variable-length video and require closely time-aligned multimodal inputs, we deliberately confine every model query to this same fixed 4-second, 12-frame window, with sensor-delta values computed over the identical interval (Figure~\ref{fig:data_preprocessing}; Section~\ref{training}); video and sensor streams are therefore synchronized by construction, and the VLM is never asked to reason over an open-ended or unaligned sequence.

\subsubsection{Sensor Data Processing}\label{sec:sensor}
\begin{figure}[h]
    \centering
    \includegraphics[scale=0.5]{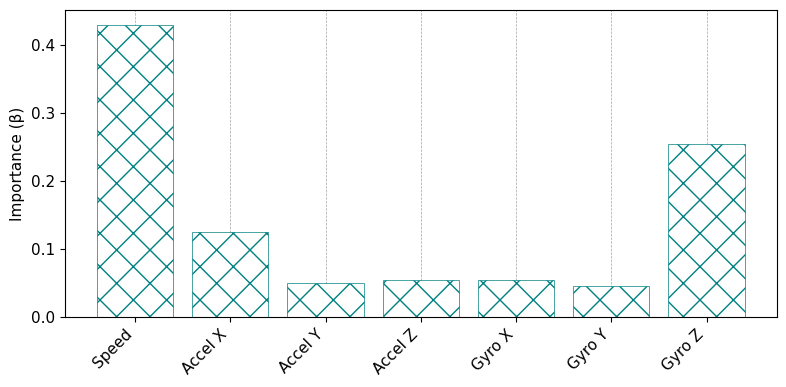}
    \caption{Importance coefficients for the seven candidate sensor channels, highlighting speed, accelerometer X, and gyroscope Z as the most informative.}
    \label{fig:importance}
\end{figure}

 The CoVLA and our field collected datasets each record a comprehensive set of six‐axis IMU and CAN‐bus channels, yet not every measurement contributes meaningfully to maneuver recognition. Nevertheless, feeding all features blindly into the VLM would negatively affect the understanding of the model; therefore, we focus on the most relevant and non-redundant variables to reduce both dimensionality and runtime with minimal information loss \cite{barbieri2024analysis}. In order to select these crucial variables from our data, we trained an interpretable multi‐variable LSTM (IMV‐LSTM)~\cite{guo2019exploring} on the full set of speed, three‐axis accelerations, and three‐axis gyroscope readings to predict the ground‐truth maneuver labels (e.g., left turn, right turn, straight, stop). IMV‐LSTM yields an importance coefficient ($\beta$) for every sensor channel, allowing us to understand and rank which features play a significant role in the classification accuracy. Hence, once training converged achieving 96\% F1-score, we selected the three most informative sensor channels: vehicle speed (derived from wheel odometry via CAN-bus), gyroscope Z-axis (yaw rate from IMU), and accelerometer X-axis (longitudinal acceleration from IMU), then we discarded all others, as shown in Figure~\ref{fig:importance}.

\subsection{Spoofing Attack Scenarios}\label{sec:attack}

In this work, we consider an adversary capable of spoofing GNSS, altering the reported position, heading, or motion state. We also assume the attacker does not tamper with cameras and in-vehicle sensors (speed, longitudinal acceleration, yaw rate). Our defense validates GNSS at the behavior level. For each short time window, the system compares the maneuver or motion state implied by GNSS with the behavior inferred from synchronized camera frames and vehicle-side sensor readings. A window is flagged when these two behavior estimates disagree in a consistent way, for example, if GNSS implies a left turn while the camera and motion signals indicate a right turn or straight movement. Within this threat space, we select attacks whose spoofed GNSS trace changes the behavior inferred by the navigation system, either through turn-direction manipulation or moving/stationary-state manipulation, aligning with prior works ~\cite{dasgupta2022sensor, zeng2018all}. Following the GNSS attack classification framework of Van Der Merwe et al.~\cite{van2018classification}, we consider attacks that manipulate spatial-temporal positioning through directional manipulation, position freezing, and false motion injection:

\begin{itemize}
 \item[-] \textbf{Wrong-Turn Attack}: The attacker intercepts GNSS signals before turning maneuvers and manipulates the positional data to indicate an incorrect turn direction. For example, spoofed data may suggest a left turn when the vehicle should turn right.
    
  \item[-] \textbf{Overshoot Attack}: The attacker replays a fixed GNSS position while the vehicle moves forward, creating a "frozen" location effect. The navigation system perceives the vehicle as stationary although it is moving, which can lead to route miscalculations when approaching intersections or decision points.

  \item[-] \textbf{Stop Attack}: While the vehicle is stationary, the attacker broadcasts false GNSS coordinates that simulate vehicle movement. This mismatch causes the navigation system to believe the vehicle is moving when it is actually stopped, potentially triggering inappropriate autonomous responses.
\end{itemize}

Having prepared synchronized 4-s window-clips (12 frames at 3 fps) along with their corresponding sensor data, we have a consistent temporal grid and a sensor reference for maneuver labels (straight, stopped, turning left, turning right). We now use these prepared clips and labels to define and instantiate the spoofing attacks. Hence, we generate three attack types from our own driving dataset (Evaluation dataset) modifying only the GNSS coordinates, and leaving window-clips and in-vehicle sensors unchanged. The key challenge here is how to ensure that generated spoofed trajectories are actually aligned with the geometry of the roads. If we simply shift coordinates in the GNSS traces without considering the road network, the resulting paths would deviate from actual roads and become easily detectable if the defender is using map data as backup. To address this, we apply road-matching techniques that align spoofed trajectories with legitimate road geometry, making the attacks realistic and harder to distinguish from authentic navigation data.

\textbf{Wrong-Turn Attack Generation:}\label{sec:wrong_turn} 
Given our manually labeled dataset of driving maneuvers, we begin by identifying all segments corresponding to turning actions, specifically left and right turns. We extract the \emph{window-clips} of each identified turn, including preceding points that capture the natural transition into the maneuver. 

The core simulation process involves mirroring the vehicle's positional data along the longitude axis, effectively swapping left turns with right turns and vice versa. However, naive mirroring of raw positional data often produces nonexistent or geometrically inconsistent routes that would be easily detectable. To address this limitation, we employ a road-matching process using the OSRM (Open Source Routing Machine) service~\cite{osrm_software}. This process "snaps" the mirrored trajectory onto the nearest feasible roads, ensuring that spoofed paths remain realistic and consistent with the underlying road network topology. This process followed by manual verification to ensure the quality of the attack.

\textbf{Vehicle Motion State Attack Generation:}
Both Stop and Overshoot attacks manipulate the relationship between actual vehicle motion and reported GNSS position, but in opposite directions.

In the \textit{Stop Attack simulation}, we first identify all dataset segments where the vehicle is confirmed stationary based on ground-truth labels. Within these stationary windows, we generate spoofed GNSS trajectories by simulating a controlled linear drift, typically shifting the latitude northward by approximately 20 meters over the stop duration while maintaining constant longitude. This creates the realistic illusion of slow vehicle movement despite the actual stationary state. To ensure spatial consistency, these spoofed coordinates are refined using OSRM road-snapping, guaranteeing that the synthetic trajectory conforms to drivable road geometry and avoids implausible deviations. 

\begin{figure}[h]
  \centering

    \begin{subfigure}[b]{0.45\textwidth}
    \centering
    \includegraphics[height=5cm,keepaspectratio]{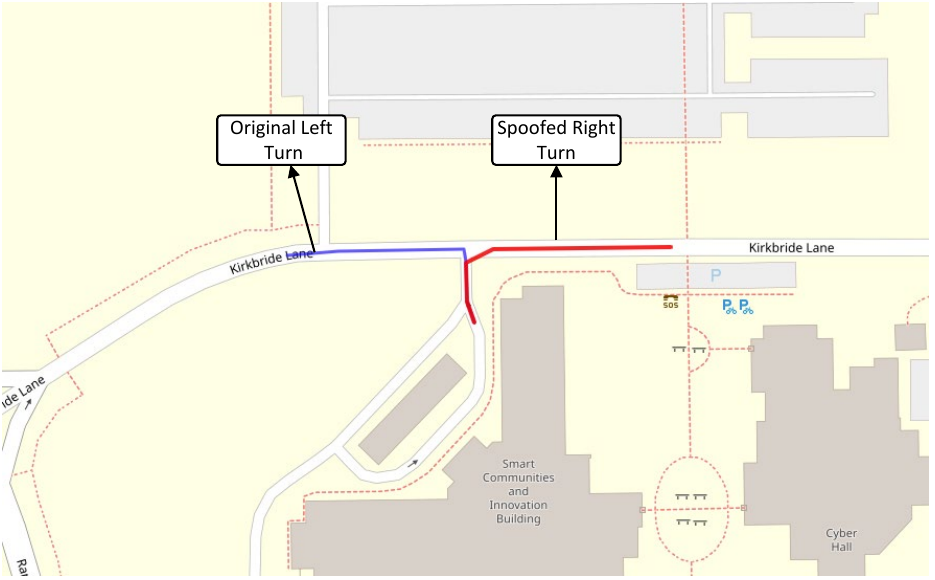}
    \caption{\centering Wrong turn: Right turn}     \label{fig:modal_image}
  \end{subfigure}
  \hfill
  \begin{subfigure}[b]{0.45\textwidth}
    \centering
    \includegraphics[height=5cm,keepaspectratio]{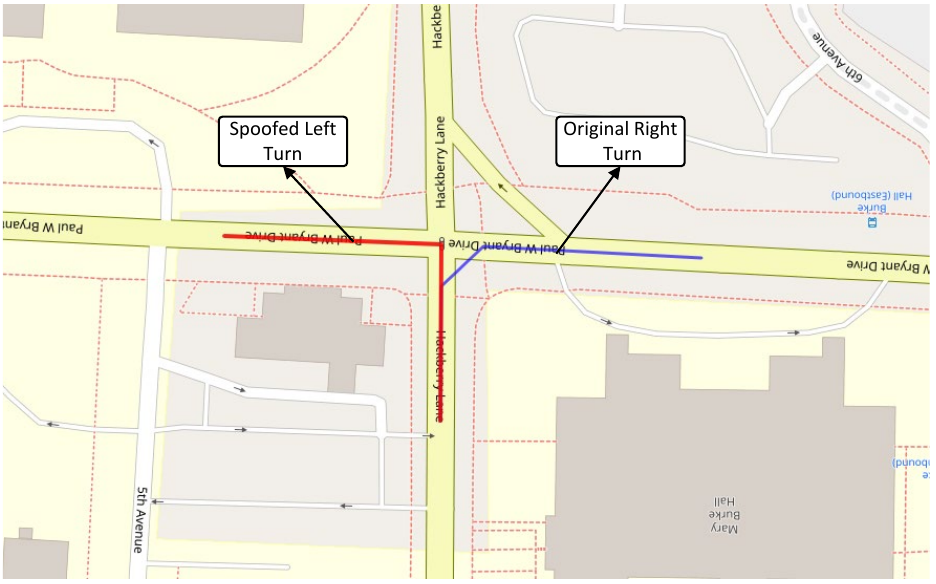}
    \caption{\centering 
    Wrong turn: Left turn}
    \label{fig:modal_ocr}
  \end{subfigure}

  \vspace{1em}

    \begin{subfigure}[b]{0.45\textwidth}
    \centering
    \includegraphics[height=5cm,keepaspectratio]{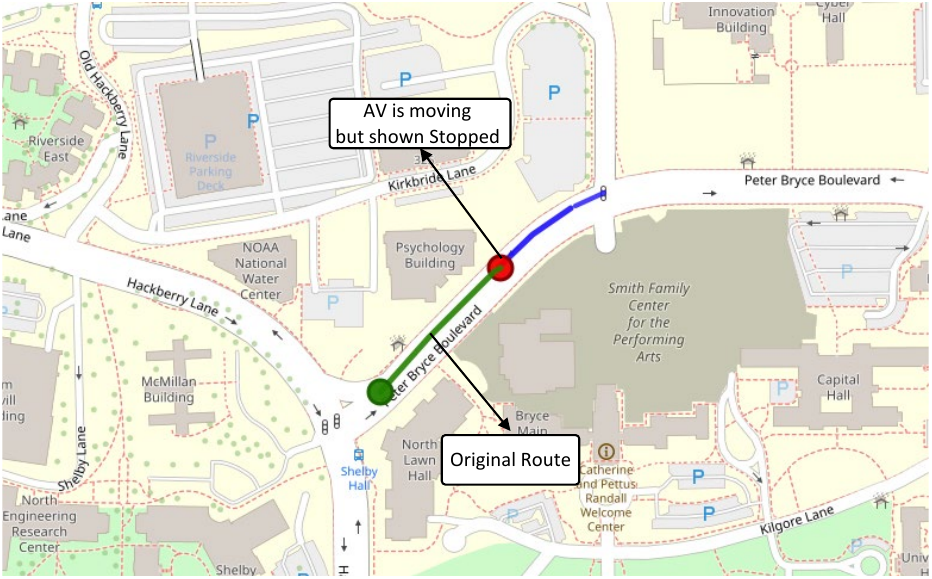}
    \caption{\centering Overshoot Attack}
    \label{fig:modal_transcript}
  \end{subfigure}
  \hfill
  \begin{subfigure}[b]{0.45\textwidth}
    \centering
    \includegraphics[height=5cm,keepaspectratio]{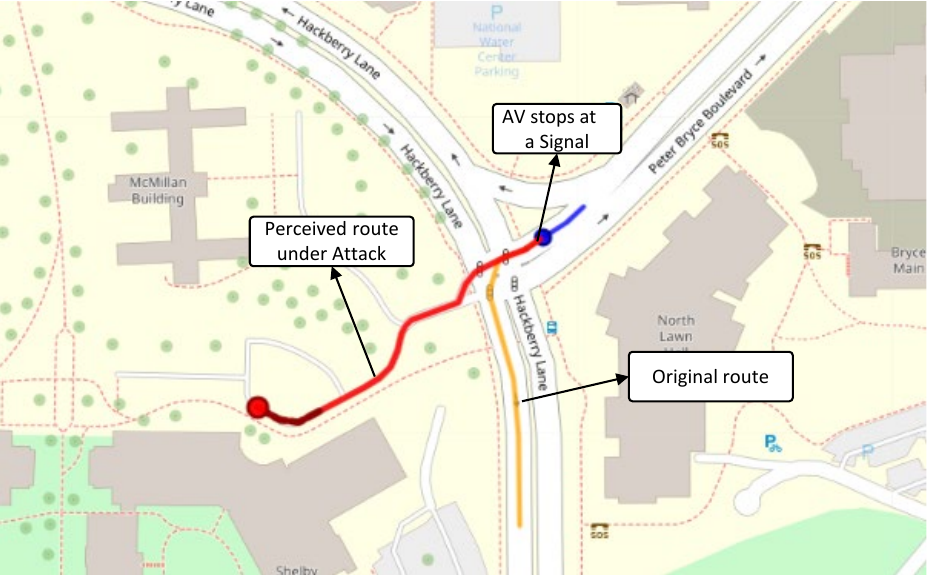}
    \caption{\centering Stop Attack}
    \label{fig:modal_imageocr}
  \end{subfigure}

  \caption{Generated GNSS spoofing attack models}
  \label{fig:modalities}
\end{figure}

For \textit{Overshoot Attack simulation} we target segments where the vehicle moves straight and freeze the reported GNSS position to the initial coordinate of the segment. This creates a constant position output that contradicts the vehicle's actual forward motion. It is important to note that, we preserve all original timestamps to maintain temporal alignment between spoofed positional data and authentic sensor measurements, ensuring that the attack remains temporally consistent while creating the dangerous illusion of a stationary vehicle during active movement. Figure \ref{fig:modalities} illustrates examples of generated attack trajectories, showing how spoofed paths (red) deviate from original routes (blue/green) while remaining aligned with the underlying road network.

\textbf{Attack Scope:} the attack generation  in this paper operates at the post-receiver trajectory level by manipulating GNSS coordinate outputs rather than generating authentic radio-frequency (RF) spoofing signals~\cite{kriezis2025real}. Real RF-level spoofing attacks (signal-level attacks) involve broadcasting counterfeit satellite signals with precise power control, carrier phase alignment, and coordination across multiple satellite channels, and interact with receiver tracking loops in ways that trajectory manipulation cannot capture. VLM-based detection framework operates at the behavioral and decision-making level rather than analyzing RF signal characteristics. Hence, RF-level signal generation and transmission mechanisms are beyond the scope of this work.

\begin{figure}[h]
\vspace{10pt}
    \centering
    \includegraphics[scale=0.53]{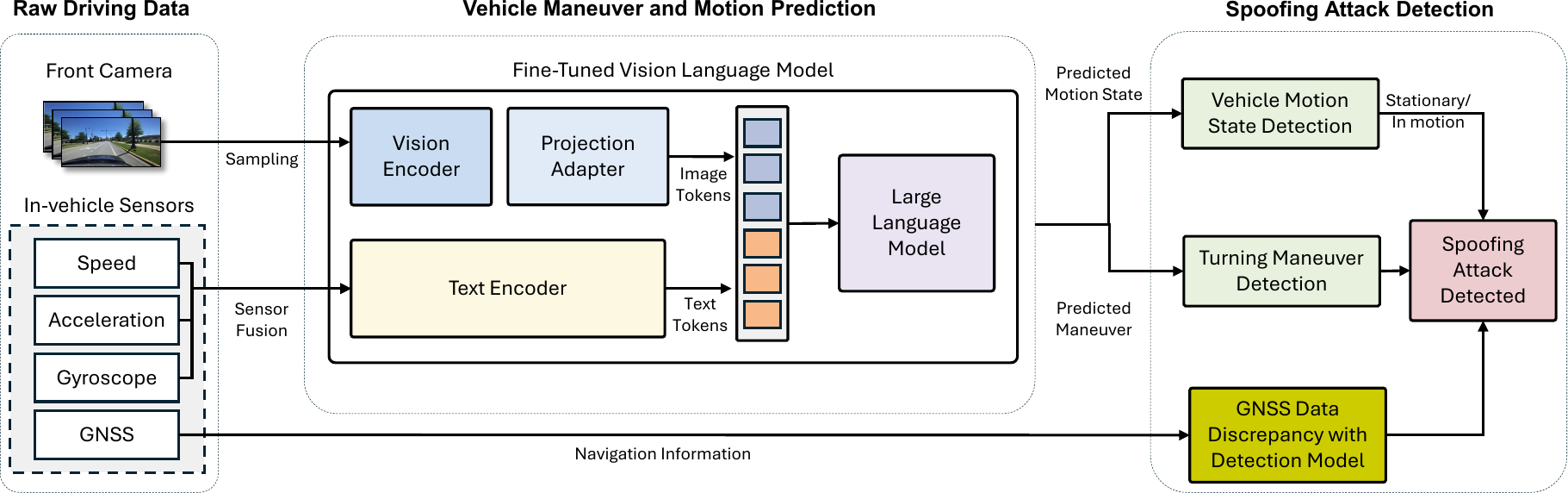}
    \caption{System overview of the proposed framework. The fine-tuned VLM receives front-camera frames and  vehicle sensor data as input and outputs the predicted vehicle status, which is passed to the turning-maneuver and vehicle-motion-state detection modules for comparison with GNSS-derived behavior.}
    \label{fig:overview}

\end{figure}
\section{VLM-Based Spoofing Detection framework}

\subsection{Overview}

Figure \ref{fig:overview} depicts the overall architecture of our VLM-based GNSS spoofing detection framework. First, the system processes synchronized front-camera data and in-vehicle sensor data, as detailed in Section \ref{sec:sensor} into structured 4-second window-clips. These preprocessed inputs are then fed into our fine-tuned Vision-Language Model, which has been trained through a three-stage process to understand the relationship between visual cues and vehicle dynamics: (i) visual cues that reveal true vehicle behavior, (ii) vehicle dynamics that reflect actual motion, and (iii) the critical relationship between them. The fine-tuned VLM predicts vehicle maneuvers (straight, turning left, turning right, or stopped) based on this multimodal observation, while simultaneously, a parallel GNSS-based system derives expected maneuvers from satellite positioning data. The framework then channels these dual predictions into two specialized detection modules: Vehicle Motion State Detection compares actual vehicle motion against GNSS-reported position changes to detect overshoot and stop attacks, and Turning Maneuver Detection evaluates turning directions between VLM predictions and GNSS-derived navigation commands to identify wrong-turn attacks. When either module detects significant disagreements, the system triggers spoofing alerts. To achieve real-time performance, the framework employs adaptive inference optimization that reduces computational costs by running VLM analysis only when sensor data indicates actual vehicle maneuver changes are occurring, rather than processing every single frame continuously.

This design directly operationalizes the two practical limitations on VLMs raised in Section 1: the need for precise cross-modal synchronization and the limited temporal-reasoning capacity of current VLMs over long or variable-length video. Synchronization is resolved entirely in preprocessing rather than left to the VLM: every window-clip pairs twelve time-aligned frames with a sensor-delta summary computed over the same 4-second interval (Section 3.3), so the model is never asked to align raw, unsynchronized streams on its own. The temporal-reasoning limitation is addressed in a similar way: rather than processing a continuous or open-ended driving sequence, each VLM query is restricted to this same fixed 4-second window, which reduces maneuver and motion-state prediction to a short-window classification problem rather than a long-horizon video-reasoning problem. The three-stage fine-tuning process (Section 4.2) then teaches the model to use this constrained, pre-aligned input effectively, rather than relying on the kind of unconstrained temporal reasoning across long sequences that current VLMs handle unreliably

\subsection{VLM Fine-Tuning for Maneuver Prediction}\label{training}
Because our detector compares VLM-predicted maneuvers with GNSS-derived maneuvers, accurate VLM maneuver prediction is a prerequisite for the entire framework. We therefore first assess whether off-the-shelf VLMs can recognize basic driving maneuvers from synchronized video and in-vehicle sensors. However, when we evaluate two state-of-the-art pre-trained VLMs—\texttt{LLaVA-Next-video-7B} and \texttt{NVILA-8B-Video} in zero-shot mode to classify driving maneuvers from dashboard camera and sensor readings, they achieved only 23–32\% accuracy. This poor performance stems from  the fact that these pretrained models were not exposed to automotive sensor data during training. Hence, they lack an understanding of the precise temporal relationships between visual cues and vehicle dynamics, and cannot reliably interpret the raw numeric telemetry (e.g., speed, acceleration, yaw rate) that characterizes driving maneuvers. To address this limitation, we need to first solve a foundational challenge: how to represent driving data in a format the VLM can process, ensuring the VLM can jointly infer maneuvers from both camera data and in-vehicle sensors.

\textbf{Prompt Design:} A critical challenge in applying VLMs to driving maneuver prediction: how to encode that sensor data into prompts the model can process and understand. While traditional autonomous driving networks process sensor readings as raw numeric arrays alongside image sequences, our framework serializes these metrics into plain text sentences. This design choice is required by the architecture of foundational VLMs, since the large language model backbone utilizes a text tokenizer built for natural language rather than raw numbers, translating continuous sensor telemetry into text is necessary to make the vehicle's physical dynamics readable to the model without altering its underlying architecture. More importantly, converting sensor data into text creates a shared semantic space between the two modalities. Instead of forcing a network to artificially merge high-dimensional video pixels with low-dimensional numbers, both inputs are unified into a single text-and-image sequence. This common language interface allows the model to directly cross-examine what it sees in the video frames against what it reads in the text prompt, which is what enables our three-stage fine-tuning later. This approach follows the established design principles of recent autonomous driving frameworks, such as DriveGPT4~\cite{xu2024drivegpt4} and DriveVLM~\cite{tian2024drivevlm}, where text-based serialization serves as a bridge that unlocks the model's pre-trained reasoning path.

 However, listing raw sensor readings from every individual video frame would create excessively long text prompts, exhausting GPU memory and degrading model performance (multiplies self-attention FLOPs)~\cite{cao2024madtp}. To prevent this token bloat, our prompt design encodes only the net change (delta) of the sensors over each 12-frame clip. This delta-style prompt is both shorter and foregrounds what actually changed during the maneuver. The structured prompt is formulated as follows:

\begin{quote}
\textit{Using this video clip and the last 12 telemetry readings from the ego vehicle, longitudinal acceleration moved from +0.1083 to -0.0721, speed changed from +0.0081 m/s to +0.0167 m/s, and yaw changed from +0.004500 to -0.019108, describe the current driving status of the ego vehicle in this video.}
\end{quote}

 The longitudinal acceleration provides direct insight into the driver’s intended longitudinal control, clearly delineating accelerative or decelerative intentions. Similarly, the speed sensor not only complements the accelerator input but captures nuanced velocity adjustments that may reflect subtle changes in maneuvering intent. Finally, yaw rate, extracted from the gyroscopic Z-axis, is crucial for discriminating lateral control actions (e.g., lane changing, turning) from straight-line cruising.

\begin{figure}[h]
    \centering
     \includegraphics[scale=0.6]{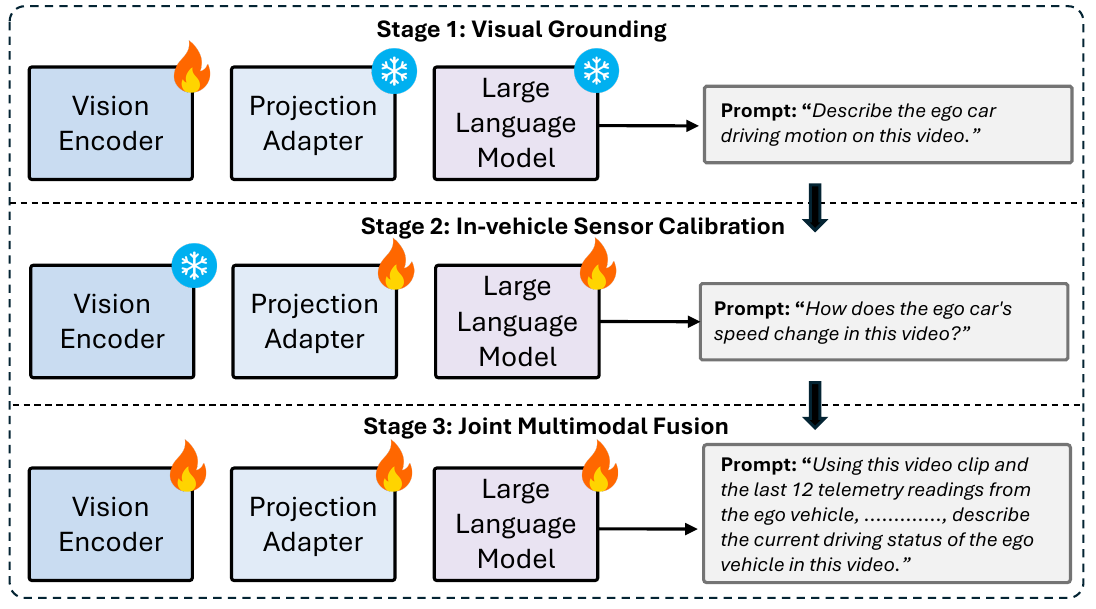}
    \caption{Three-stage fine-tuning pipeline: (1) Visual Grounding – train only the vision encoder on video clips sequences while freezing the adapter and LLM; (2) Sensor Calibration – freeze the vision encoder and fine‐tune the adapter and LLM on $\Delta$-style prompts of accelerator, speed, and yaw changes; (3) Joint Multimodal Fusion – unfreeze all components and jointly fine‐tune on concatenated image embeddings, sensor‐delta sentences for balanced vision–sensor reasoning.}
    \label{fig:stages}
\end{figure}

\textbf{Fine-Turning Process:} \label{sec:fine_tuning}
A naïve one–shot fine-tune that feeds all images and sensor tokens into the VLM at once quickly collapses in practice. For example, high-variance gradients from raw pixels disturb the language weights, corrupting their ability to parse numeric telemetry. Also, the network focus on any cue such as speed spikes, without truly coupling vision (frames) to vehicle dynamics (sensors). To solve this issue, we adopt three-stage tuning in which each stage isolates and stabilizes one modality before allowing the next to learn, as illustrated in Figure \ref{fig:stages}.

On the first stage, we train the vision encoder (3 epochs) while keeping the parameters of other modules frozen, such as the adapter and the LLM. This enables the vision encoder to project raw image features directly into the LLM’s semantic embedding space.  Each training sample in this stage supplies the frames from Section \ref{sec:video} together with a fixed query ({\it ``Describe the ego car
driving motion on this video?''}). The query is tokenised once, cached, and passed through the frozen
language stack solely to obtain a text embedding; it is not updated.

On the second stage, we aim to teach the adapter and LLM to understand our vehicle’s sensor readings within the same semantic space that Stage 1 established for visual inputs. To do this, we freeze the vision encoder—so that all key-frame representations learned in the previous remain fixed—and unfreeze the adapter together with the large language model for two epochs. In each iteration, the model takes: \textit{(1)} the frozen visual embeddings, \textit{(2)} a textual prompt query to predict the net speed of the ego car in that video clip, and (3) the ground truth caption (e.g., “speed decreased steadily from 32.17 ft/s to 17.52 ft/s,”). This process grounds vision features in the same semantic space as our numeric telemetry, forcing the adapter to translate frozen visual embeddings into numeric tokens.

On the last stage, we unfreeze all parameter—the vision encoder, adapter, and the full language model—and present both modalities together for  10-epoch joint fine-tune. Each data entry is ordered as:

\[
\makebox[\linewidth]{  $\displaystyle
    \underbrace{{\tt <IMG>} \,\mathbf{z}_{1},\dots,\mathbf{z}_{12}}
                _{\text{key‑frame embeddings}}
    \;|\;
    \underbrace{\text{sensor‑delta sentence}}
                _{\Delta\text{accelerator},\,\Delta\text{speed},\,\Delta\text{yaw}}
    \;|\;
    \underbrace{\text{maneuver caption}}
                _{\textit{``turning left''}}  $}
\]
\hfill\break

where the {\tt <IMG>} tag is internally replaced by the adapter’s image
embedding~$\mathbf{z}_i$.  This three-stage fine-tuning grounds visual semantics, secures sensor-numeric understanding, and then fuses the two streams so that neither modality dominates, enabling the VLM to reason jointly over images and telemetry with balanced attention to both.

A critical outcome of this formulation is cross-regional generalization. Three aspects of our design contribute to the model's ability to transfer learning across fundamentally different road environments (e.g., from Tokyo to Tuscaloosa). First, our delta-style sensor prompts encode only the net change in speed, longitudinal acceleration, and yaw rate over each 4-second clip. These quantities are governed by vehicle dynamics: a left turn produces a characteristic negative yaw rate delta and lateral acceleration regardless of whether the vehicle is in Tokyo or Alabama. Because the input features reflect the physics of vehicle motion rather than local road geometry, signage, or lane markings, they naturally transfer across environments. Second, the first stage of the fine-tuning process trains the vision encoder with the prompt \textit{“Describe the ego car driving motion on this video?”}This directs the visual module toward temporal motion cues, such as how the visual scene shifts across consecutive frames during turning, stopping, or straight movement, rather than toward static scene elements such as signs, markings, or lane geometry.  When a vehicle turns left, surrounding objects shift rightward in the camera view regardless of what those objects look like; these motion cues are inherently domain-invariant.Third, the VLM backbones used in this study, LLaVA-Next-video-7B and NVILA-8B-Video, are initialized from large-scale pretrained vision-language models with broad visual-semantic knowledge. Our staged fine-tuning preserves this pretrained general visual representation while adding task-specific driving-behavior understanding. In other words, our three-stage fine-tuning teaches the model one specific skill at a time. In Stage 1, we lock the model to only train its vision system to track driving motion. In Stage 2, we lock the vision system and only teach the model language encoder how to read numerical sensor changes (speed and yaw). Stage 3 jointly fine-tunes both modalities for maneuver and motion-state prediction. This staged design reduces the risk that the model learns only region-specific visual shortcuts from CoVLA. 

\subsection{Attack Detection Logic}\label{sec:attack}
In this research, the GNSS spoofing detection approach relies on identifying inconsistencies between the derived maneuvers from the GNSS and the predicted maneuvers from the VLM that integrates fused visual data (camera) and inertial data (sensors). Specifically, our framework exploits discrepancies in turning maneuver and vehicle motion state to detect spoofing attacks.

\subsubsection{Turning Maneuver Detection}

First, the fine-tuned VLM takes as input the selected clip (12 frames) together with the associated net change in the vehicle sensor readings (speed, acceleration X-axis, yaw-rate Z-axis) computed over that same temporal window. From these multimodal inputs, the VLM generates a maneuver classification among four possible categories: \textit{going straight}, \textit{turning left}, \textit{turning right}, or \textit{stopped}. Formally, for each \emph{4\,s window-clip}.

Meanwhile, the maneuver derived from the GNSS can be predicted using a standard GNSS navigation system, inspired by \cite{dasgupta2022sensor}, by analyzing the GNSS location data in combination with the road link identification obtained from a built-in map \cite{zhao2018analysis}. As an AV navigates, the GNSS location data continuously updates the vehicle’s position, allowing the system to identify the current road link ID. When the vehicle executes a turn, the road link ID changes. We monitored consecutive link IDs. A turn event flagged whenever the current link ID differs from the previous one. To classify the direction of detected turns, we selected the two GNSS samples immediately before and after the link transition and computed their bearing angles:

\begin{equation}
\theta 
\;=\;
\operatorname{atan2}\!\bigl(
   \sin\Delta\lambda\,\cos\varphi_{2},\;
   \cos\varphi_{1}\,\sin\varphi_{2}
   -\sin\varphi_{1}\,\cos\varphi_{2}\,\cos\Delta\lambda
\bigr)
\end{equation}
\hfill\break
where \((\varphi_{1},\lambda_{1})\) and \((\varphi_{2},\lambda_{2})\) are the latitude/longitude pairs and $\Delta\lambda = \lambda_{2}-\lambda_{1}$. We then compute the clockwise angle difference as:
\hfill\break
\begin{equation}
\Delta\theta \;=\;(\theta_{2}-\theta_{1}+360^\circ)\bmod 360^\circ.
\end{equation}
\hfill\break
If $\Delta\theta$ falls between $0^\circ$ and $180^\circ$, the maneuver is classified as a right turn; if it falls between $180^\circ$ and $360^\circ$, classifies as a left turn.

\subsubsection{Turning–Maneuver Agreement Rule}
For each 4\,s window-clip $W_t=\{I_{t-11},\ldots,I_t\}$ (12 frames at 3\,FPS), we evaluate turning consistency using a subset of the most recent $N_k=8$ \emph{keyframes} $K_t=\{I_{t-7},\ldots,I_t\}$ to emphasize the immediate pre-/in-turn context and reduce latency. For each keyframe $I_i\in K_t$, the VLM outputs a per-frame maneuver $\hat{y}_i \in \{\text{left},\text{right},\text{straight},\text{stopped}\}$ given the paired signals. Let $y^\ast \in \{\text{left},\text{right}\}$ denote the turn expected from the GNSS/route planner.

We declare the segment \emph{non-spoofed} (i.e., not spoofed) if at least $T=4$ keyframes agree with the expected maneuver:
\hfill\break
\[
\sum_{i \in K_t} \mathbf{1}[\hat{y}_i = y^\ast] \;\ge\; T,
\]
\hfill\break
and \emph{spoofed} otherwise. In other words, at least half of the $N_k=8$ keyframes must match the planned turn. For example, if the route expects a \emph{right} turn but only $1$ of the $8$ keyframes is classified as \emph{right} ($1 < T{=}4$), we flag the segment as spoofed.

\subsubsection{Vehicle Motion State Detection}

Given that our VLM was trained to detect more than just the 
turns of an AV, we extended our model to identify other 
spoofing scenarios that exploit vehicle motion state, such as 
\textit{overshoot} or \textit{stop} attacks. Unlike turning maneuver detection, which is triggered only when a road link ID transition is detected, vehicle motion state detection runs continuously on every 4-second window. Specifically, we use two consecutive GNSS position samples and compute the geographical distance between these points to infer the vehicle’s motion state. For each sample, we have a latitude-longitude pair $(\varphi, \lambda)$. To accurately measure the distance traversed between two consecutive samples, we employ the Haversine formula, defined as follows:

\begin{equation}
  d = 2 R \arcsin\!\Biggl(\sqrt{ 
      \sin^2\!\Bigl(\frac{\Delta\varphi}{2}\Bigr)
      + \cos\varphi_1\;\cos\varphi_2\;\sin^2\!\Bigl(\frac{\Delta\lambda}{2}\Bigr)
    }\Biggr)
\end{equation}
\hfill\break
where:

\[
  \Delta\varphi = \varphi_2 - \varphi_1,\quad
  \Delta\lambda = \lambda_2 - \lambda_1,\quad
  R \approx \SI{6371}{\kilo\metre}\ \text{(Earth's radius)}
\]

The computed distance \(d\) indicates whether the vehicle has physically moved. Dividing  \(d\) by the known sampling interval \(\Delta t\) gives a speed
\hfill\break
\begin{equation}
  v = \frac{d}{\Delta t}, 
  \quad
  \Delta t = \frac{1}{f_s}
\end{equation}
\hfill\break
where \(f_s\) is the GNSS update rate (in Hz), i.e.\ the number of position fixes per second (e.g.\ if \(f_s=3\) Hz, then \(\Delta t=1/3\) s). If this speed \(v\) exceeds the threshold 2 $m/s$ within the selected window, the vehicle is classified as \emph{moving}; otherwise, it is classified as \emph{stationary}. This threshold of 2 $m/s$ ($\approx$ 7.2 km/h) lies within the 5–8 km/h boundary widely used by traffic sensing practice, which separates very low-speed/stopped vehicles from vehicles in motion \cite{klein2006traffic}.  For each 4-second window, the GNSS-derived motion state is first estimated from consecutive GNSS positions and compared with the motion state inferred from in-vehicle speed. If the two motion-state estimates disagree persistently across consecutive windows, the VLM is invoked to provide semantic confirmation using synchronized camera frames and vehicle-side sensor readings.

\subsection{Inference Optimization}\label{sec:adaptive}
Executing VLM inference on every temporal window presents significant computational challenges in real-world deployment scenarios since they demand substantial GPU memory and computational resources. To address this bottleneck while maintaining detection accuracy, we developed an adaptive prediction strategy that reduces VLM invocations without compromising spoofing detection performance.

\begin{figure}[h]

    \centering
     \includegraphics[scale=0.67]{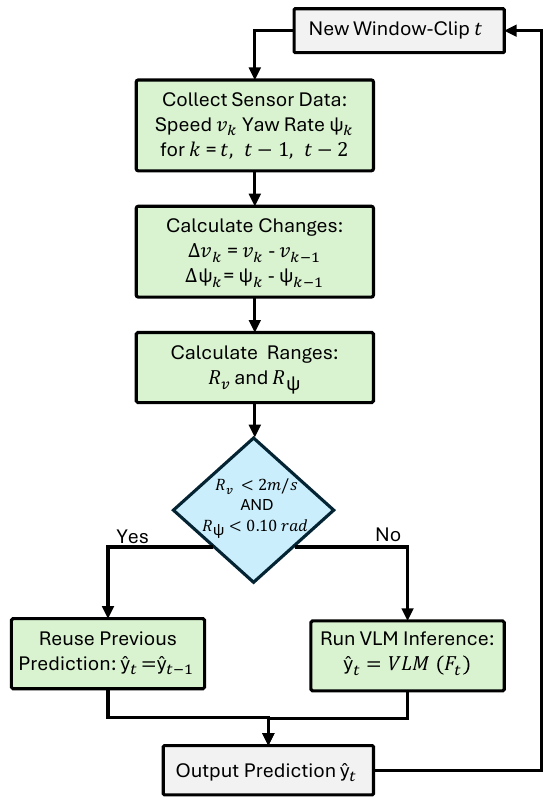}
    \caption{Real-time adaptive VLM inference workflow}
    \label{fig:flowchart}
    \vspace{-30pt}
\end{figure}

Based on the observation that sensor measurements (speed, acceleration, and yaw rate) remain within predictable ranges across consecutive window‑clips, the underlying driving maneuver is likely unchanged. For example, if a vehicle maintains nearly constant speed and minimal yaw rate variation over several windows, it is reasonable to infer that the vehicle is continuing the same maneuver, such as driving straight or remaining stopped. Therefore, we calculate the change in sensor readings for speed and gyroscope over a sliding window of the last three temporal segments. Let \(v_k\) and \(\psi_k\) be the speed and yaw‐rate measurements at segment \(k\), and define:

\begin{equation}
\Delta v_k = v_k - v_{k-1}, 
\quad
\Delta \psi_k = \psi_k - \psi_{k-1},
\qquad
k \in \{t-2,\,t-1,\,t\}
\end{equation}
\hfill\break
Then we compute the ranges of change in these sensors:

\begin{align} R_v(t) &= \max_{k\in\{t-2,t-1,t\}}\Delta v_k - \min_{k\in\{t-2,t-1,t\}}\Delta v_k, \\ R_\psi(t) &= \max_{k\in\{t-2,t-1,t\}}\Delta \psi_k - \min_{k\in\{t-2,t-1,t\}}\Delta \psi_k \end{align}
\hfill\break
After that we compare the change based on a fixed thresholds: \(\tau_v = 2.0\,\mathrm{m/s}\) and \(\tau_\psi = 0.1\,\mathrm{rad}\), our causal decision rule is:
\hfill\break
\begin{equation}
\hat y_t =
\begin{cases}
\hat y_{t-1}, & R_v(t) < \tau_v \;\wedge\; R_\psi(t) < \tau_\psi,\\[6pt]
\mathrm{VLM}(F_t), & \text{otherwise},
\end{cases}
\end{equation}
\hfill\break
where \(F_t\) denotes the current \emph{4\,s window-clip}. So if either range exceeds its threshold, we trigger the VLM inference on the current window sequence to predict the maneuver transition. Figure~\ref{fig:flowchart} illustrates the complete adaptive inference workflow, which leverages sensor-based gating to determine when VLM inference is necessary versus when previous predictions can be reused.

The adaptive inference policy operates differently for turning-maneuver detection and vehicle-motion-state detection. For turning-maneuver detection, the VLM is invoked only when kinematic changes in vehicle speed or yaw rate exceed predefined thresholds, indicating that the vehicle may be entering a new maneuver. This kinematic gating mechanism allows the framework to selectively evaluate whether the GNSS-derived turn direction agrees with the driving behavior inferred from camera and vehicle-side sensor evidence, safely skipping VLM calls during stable, straight-line segments. For motion-state attacks, including overshoot and stop attacks, the framework uses a separate motion-state disagreement trigger. On every window, the Haversine-derived GNSS speed is mathematically calculated and compared directly against the in-vehicle sensor speed.  If GNSS-derived speed indicates that the vehicle is stationary while the in-vehicle speed indicates movement, or if GNSS-derived speed indicates movement while vehicle-side evidence indicates that the vehicle is stopped, the VLM is invoked to confirm the motion state using synchronized camera frames and sensor telemetry. The attack is detected because the frozen GNSS position creates a persistent contradiction between the GNSS-derived motion state and the vehicle’s observed motion state.

It is also important to note that the adaptive gate is introduced to reduce the computational cost of continuously invoking the VLM, rather than a claim that VLMs are inherently cheaper or more lightweight than traditional VIO frameworks. It is also not intended to assume that spoofing attacks only occur when speed or yaw-rate changes exceed the thresholds. When speed and yaw-rate remain stable, the system skips the VLM call because the current maneuver or motion state is likely unchanged. However, this creates a trade-off: a slow-drift or carry-off attack that preserves nearly constant speed and heading may keep the event trigger inactive and delay VLM re-evaluation. To reduce this risk, the adaptive policy can be implemented as a hybrid event‑ and time‑triggered rule. In practice, this means the VLM is called either when the speed or yaw‑rate trigger is active, and a time‑based rule forces a new VLM call every $K$ windows, preventing the gate from relying on constant speed and heading for too long. This periodic re-evaluation prevents the detector from relying indefinitely on unchanged vehicle dynamics. At each forced VLM call, the system re-estimates the current maneuver or motion state from synchronized camera frames and vehicle-side sensor readings and compares this estimate with the GNSS-derived behavior. If a slow‑drift attack eventually leads to a noticeable change in driving behavior, our model is able to detect it.

\section{Evaluation Results}

\begin{figure}[h]

    \centering
     \includegraphics[scale=0.45]{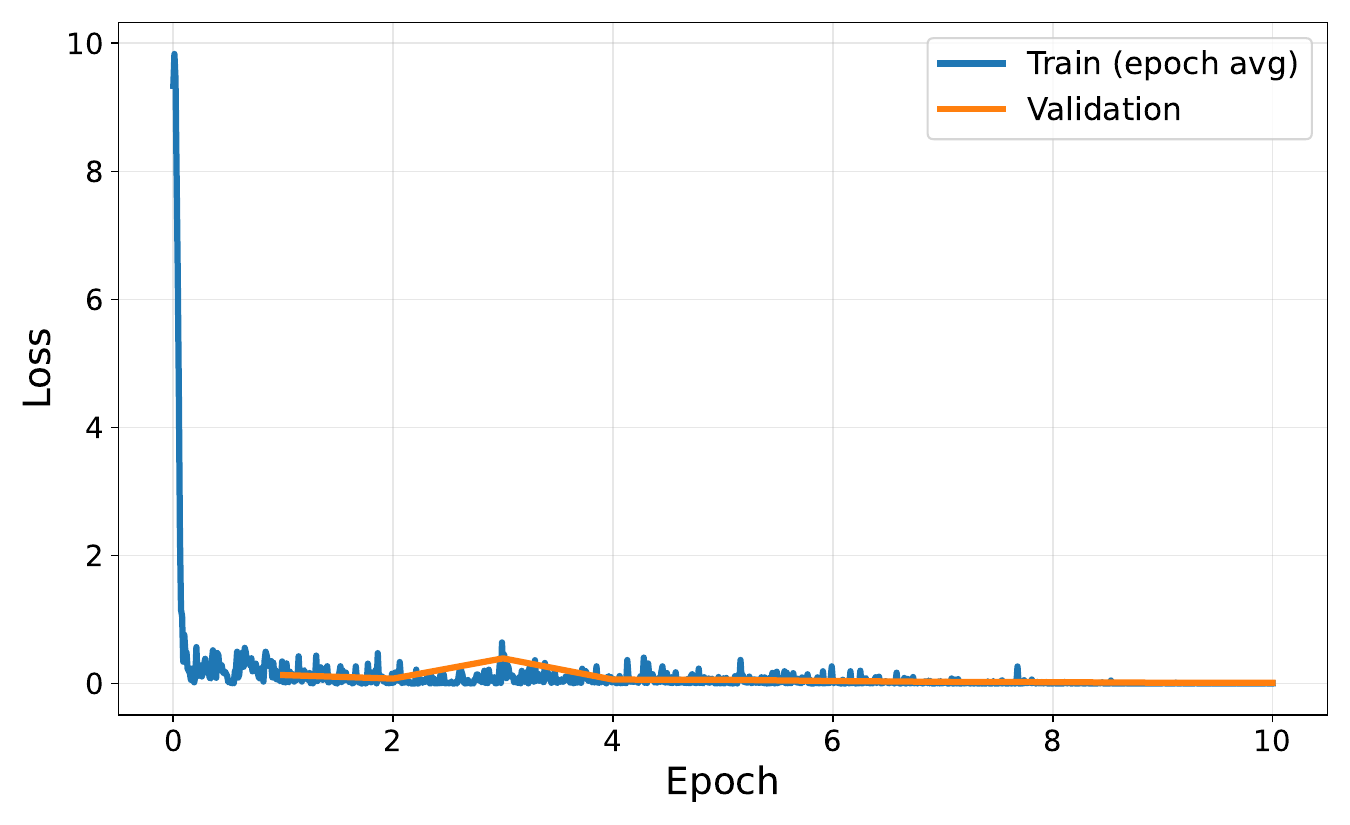}
     \vspace{-15pt}
    \caption{Training and validation loss during fine-tuning}
    \vspace{-10pt}
    \label{fig:training_loss}
 
\end{figure}

 All three stages of the VLM fine tuning in Section \ref{training} are trained only on CoVLA-derived window clips. We used the processed dataset from Section \ref{sec:data}, which yields 1,000 labeled window clips across four maneuvers; we use 820 clips for training and 180 clips as an internal CoVLA test set. These clips consist entirely of attack-free driving data, as the VLM is trained exclusively to learn normal baseline behavior rather than explicit spoofing attack patterns. Figure~\ref{fig:training_loss} shows stable convergence throughout training  on the last stage of fine tuning with no late epoch divergence. For validation, we used our Tuscaloosa validation dataset, which also follows the identical preprocessing of training data (3\,fps, 4\,s window‑clips, slide \(=1\)). All experiments ran on a server equipped with a 48‑core CPU, two NVIDIA A100-80GB GPUs, and Ubuntu 20.04LTS. It is important to note that, all results reported in this section are evaluated on Tuscaloosa validation dataset.

\subsection{VLM Baseline Evaluation} The evaluation began with assessing the pre-trained VLMs performance (\texttt{LLaVA-Next-video-7B} and \texttt{NVILA-8B-Video}) on zero-shot tasks using Tuscaloosa validation dataset. We then evaluate the proposed fine tuned model on the same validation set. In addition, to demonstrate the importance of visual information, we also evaluated the fine-tuned model using only the text and sensor tokens only without window-clips. From Table~\ref{table:llava7b_performance} reports overall accuracy on the Tuscaloosa validation dataset, and we can see that the substantial performance improvement from a baseline accuracy of 23-32\% to a fine-tuned 95\% indicates that VLMs were able to learn driving maneuver patterns from in-vehicle sensor and visual data. Notably, this 95\% accuracy on Tuscaloosa data confirms that the model learned physics-based motion patterns (sensor deltas and visual scene displacement) rather than region-specific visual features road geometry.  When windowed video inputs are removed, leaving only sensor data, accuracy drops to 0.82–0.86, demonstrating that fusing  between the modality in our model is robust. This highlights temporal visual information with sensor readings is essential for precise maneuver detection. In addition to overall accuracy, we evaluate per-class performance, as shown in Table \ref{tab:perclass}.

\begin{table}[htb]
\centering

  \caption{Overall performance under different training setups: (“Zero shot” refers to the pre trained VLMs without any fine tuning. “Fine tuned” refers to the three stage regimen above. “Without visual input” is an ablation that removes the windowed video and feeds only sensor tokens.)}
  \label{table:llava7b_performance}
   \resizebox{!}{1.6cm}{
    \begin{tblr}{
      colspec={X[2,l]|X[c]X[c]X[c]X[c]|X[c]X[c]X[c]X[c]},
      row{even} = {white,font=\small},
      row{odd} = {bg=black!10,font=\small},
      row{1} = {bg=black!20,font=\bfseries\small},
      hline{1,3,Z} = {1pt,solid,black!60},
      hline{2} = {solid,black!40},
      vline{2,6} = {1pt,solid,black!60},
      rowsep=3pt
    }
    \SetCell[r=2]{c} \textbf{Setup} 
      & \SetCell[c=4]{c} \textbf{\texttt{LLaVA-Next-video-7B}} 
      & & & 
      & \SetCell[c=4]{c} \textbf{\texttt{NVILA-8B-Video}} 
      & & & \\ 
    & \textbf{Accuracy} & \textbf{Precision} & \textbf{Recall} & \textbf{F1}
    & \textbf{Accuracy} & \textbf{Precision} & \textbf{Recall} & \textbf{F1} \\
    Baseline & 0.32 & 0.31 & 0.36 & 0.25
             & 0.23 & 0.25 & 0.25 & 0.21 \\
    w/o Visual Input & 0.86 & 0.88 & 0.89 & 0.88
                     & 0.82 & 0.86 & 0.85 & 0.81 \\
    Fine-tuned & \textbf{0.95} & \textbf{0.95} & \textbf{0.94} & \textbf{0.94}
               & \textbf{0.95} & \textbf{0.95} & \textbf{0.95} & \textbf{0.95} \\
    \end{tblr}  }
  \vspace{-10pt}
\end{table}

\subsection{Attack Detection Performance}
\textit{Wrong-turn Attack Detection:} Having established our attack simulation as described in Section~\ref{sec:wrong_turn}, we next evaluate our detection framework’s ability to identify these manipulated maneuvers. The modified data was fed into two separate maneuver prediction pipelines: our VLM and the GNSS-based system, as detailed in Section~\ref{sec:attack}. Specifically, the spoofed GNSS data served as input to the GNSS-based maneuver detection system, which tracked road link ID transitions and calculated bearing angle differences before and after each link change to predict the turn. On the other side, our VLM was fed the corresponding 12 front camera frames alongside the summarized sensor telemetry changes (accelerometer X-axis, speed, and yaw rate) to classify the turns.

Table~\ref{tab:wrong_turn} summarizes the predictions from both sources across all spoofed turns. It reveals a clear discrepancy: the spoofed GNSS system uniformly misclassified the turns, predicting right turns where the actual maneuver was left, and left turns where the actual maneuver was right. In contrast, both of our fine tuned VLMs (\texttt{LLaVA-Next-video-7B} and \texttt{NVILA-8B-Video}) predictions remain effectively aligned with the true maneuver directions, achieving 100\% accuracy.
\begin{table}[h]
\centering
  \caption{Breakdown of our fine-tuned VLMs performance in maneuvering tasks}
  \label{tab:perclass}
  \resizebox{!}{2.5cm}{
  \begin{tblr}{
    colspec={X[0.2,l]|X[0.3,l]|X[0.3,c]X[0.3,c]X[0.3,c]},
    width=\textwidth,
    row{even} = {white,font=\small},
    row{odd}  = {bg=black!10,font=\small},
    row{1}    = {bg=black!20,font=\bfseries\small},
    hline{Z}  = {1pt,solid,black!60},
    vline{2}  = {1pt,solid,black!60},
    rowsep=3pt
  }
   \hline
  \centering\textbf{Task} & \centering\textbf{Model} & \textbf{F1-Score} & \textbf{Precision} & \textbf{Recall} \\ \hline
  \SetCell[r=2]{l} Turning Left
    & \texttt{LLaVA-Next-video-7B} & 0.89 & 0.86 & 0.92 \\
    & \texttt{NVILA-8B-Video}      & 1.00 & 1.00 & 1.00 \\\hline
  \SetCell[r=2]{l} Turning Right
    & \texttt{LLaVA-Next-video-7B} & 1.00 & 1.00 & 1.00 \\
    & \texttt{NVILA-8B-Video}      & 1.00 & 1.00 & 1.00 \\\hline
  \SetCell[r=2]{l} Straight
    & \texttt{LLaVA-Next-video-7B} & 0.91 & 1.00 & 0.83 \\
    & \texttt{NVILA-8B-Video}      & 0.92 & 0.92 & 0.92 \\\hline
  \SetCell[r=2]{l} Stopped
    & \texttt{LLaVA-Next-video-7B} & 0.97 & 0.95 & 1.00 \\
    & \texttt{NVILA-8B-Video}      & 0.89 & 0.89 & 0.89 \\
  \end{tblr}
  }
\end{table}
\begin{table}[h]
  \centering
  \caption{Summary of wrong turn attack detection performance}
  \label{tab:wrong_turn}
  \resizebox{0.9\textwidth}{!}{    \begin{tblr}{
      colspec={X[0.3,l]X[0.2,c]X[0.4,c]X[0.4,c]X[0.3,c]},
      row{even} = {white,font=\small},
      row{odd}  = {bg=black!10,font=\small},
      row{1}    = {bg=black!20,font=\bfseries\small},
      hline{Z}  = {1pt,solid,black!60},
      rowsep=3pt
    }
     \hline
    \textbf{Actual Turn Type} & \textbf{\# of Turns} & \textbf{Predicted by GNSS} & \textbf{Predicted by our VLMs} & \textbf{Accuracy} \\ \hline
    Left Turn  & 6 & Right turns & Left turns  & 100\% \\
    Right Turn & 4 & Left turns  & Right turns & 100\% \\
    \end{tblr}
  }
\end{table}

\textit{Stop Attack Detection:}
From Table \ref{tab:stop_attack_results}, both of our fine-tuned VLMs successfully detected all instances of stop attack with 100\% accuracy. This high detection rate can be attributed to the clear discrepancy between the visual evidence (stationary vehicle captured by the camera) and sensor data (barely any speed and minimal acceleration) versus the spoofed GNSS data, which indicates movement.

\textit{Overshoot Attack Detection:}
 Across VLMs, overshoot detection accuracy is 88.33-93.4\% (53/60 and 56/60 window-clips correctly detected), as shown in Table \ref{tab:stop_attack_results}. For \texttt{LLaVA-Next-video-7B}, 3 windows were incorrectly predicted as "stopped," indicating missed detections, the remaining 1 window were classified as turning maneuvers. Considering that both turning maneuvers indicate vehicle motion (the core objective of overshoot detection), the detection rate increases to 95\%. For \texttt{NVILA‑8B‑Video}, 5 were predicted as “stopped”, while 2 were classified as turning maneuvers rounding the accuracy to (91.7\%).

\begin{table}[htb]
  \centering
  \caption{Summary of stop and overshoot attacks detection results}
  \label{tab:stop_attack_results}
  \resizebox{0.9\textwidth}{!}{    \begin{tblr}{
      colspec={X[1.3,c]|X[0.9,c]|X[c]X[c]|X[c]X[c]},
      row{even} = {white,font=\small},
      row{odd}  = {bg=black!20,font=\small},
           row{2}    = {bg=black!20,font=\bfseries\small},     row{3}    = {bg=black!10,font=\bfseries\small},
      hline{1,Z} = {1pt,solid,black!60},
      vline{2,4} = {1pt,solid,black!60},
      rowsep=3pt
    }
    
    \SetCell[r=2]{c}\textbf{Attack Type} 
      & \SetCell[r=2]{c}\textbf{Total Window-Clips}
      & \SetCell[c=2]{c}\textbf{\texttt{LLaVA-Next-video-7B}} & 
      & \SetCell[c=2]{c}\textbf{\texttt{NVILA-8B-Video}} & \\ \hline
    & & \textbf{Correctly Detected} & \textbf{Accuracy}
        & \textbf{Correctly Detected} & \textbf{Accuracy} \\\hline
    Stop Attack      & 20 & 20 & 100\%  & 20 & 100\%   \\
    Overshoot Attack & 60 & 56 & 93.4\% & 53 & 88.33\% \\
    \end{tblr}
  }
\end{table}

\subsection{Computational Performance}

The adaptive prediction strategy significantly improves computational efficiency by reducing unnecessary VLMs calls. Our lightweight sensor-based decision mechanism requires only (0.9 $\mu\mathrm{s}$ - 1.0 $\mu\mathrm{s}$) on average to determine whether VLM inference is necessary. This negligible overhead enables substantial computational savings while maintaining detection accuracy.
Out of 3,999 total predictions, our adaptive strategy required VLM inference for only 559 cases (14.0\%), achieving 86.0\% computational savings by skipping 3,440 predictions where sensor readings indicated stable driving conditions. This substantial reduction in VLM calls directly translates to energy savings and improved real-time performance.

\textbf{Real-Time Inference:} The computational efficiency of our GNSS spoofing detection framework is critical for real-time deployment in autonomous vehicles. We evaluated the timing performance of our system across 4,000 \emph{4s window-clips} (from the evaluation dataset) using our optimized adaptive prediction strategy described in the previous section. Table \ref{tab:comp_performance_summary} presents a comprehensive comparison of runtime performance between two VLM architectures. Our fine-tuned Vision Language Models demonstrate consistent inference performance with an average latency of 467.7 -  522.3ms per window-clip  with min/max 414-1{,}110ms when VLM inference is required. The standard deviation was 22 - 55ms, indicating stable computational performance across different driving maneuvers.

\textbf{Deployment Considerations:} In addition to processing time, we measured peak GPU memory allocation during VLM inference. The peak allocated memory was approximately 14 GB for LLaVA-Next-video-7B and 16 GB for NVILA-8B-Video. These values reflect the resource demand of the current uncompressed research implementation. The experiments were conducted on a development/evaluation server with a 48-core CPU and an NVIDIA A100 GPU; this platform was not intended to represent a production in-vehicle compute unit. Therefore, the reported results should be interpreted as proof-of-concept computational measurements for behavior-level spoofing detection, rather than as evidence of immediate deployability on standard automotive hardware.
The proposed adaptive inference policy partially reduces the computational burden by avoiding continuous VLM invocation. Instead of processing every camera frame, the system invokes the VLM only on selected synchronized windows where behavior validation is informative, such as maneuver-relevant windows or persistent GNSS-versus-vehicle motion-state disagreement. This reduces the number of VLM calls, but it does not eliminate the need for embedded optimization. Practical deployment would require smaller VLM backbones, quantization, pruning, distillation, and hardware-specific optimization using automotive-grade inference engines.

\begin{table}[htb]
  \centering
  \caption{Runtime performance summary}
  \label{tab:comp_performance_summary}
  \resizebox{0.9\textwidth}{!}{    \begin{tblr}{
      colspec={X[0.4,l]|X[c]|X[c]},
      row{even} = {white,font=\small},
      row{odd}  = {bg=black!10,font=\small},
      row{1}    = {bg=black!20,font=\bfseries\small},
      hline{1,Z} = {1pt,solid,black!60},
      vline{2}   = {1pt,solid,black!60},
      rowsep=3pt
    }
    \textbf{Metric} & \textbf{\texttt{LLaVA-Next-video-7B}} & \textbf{\texttt{NVILA-8B-Video}} \\
    VLM Inference    & 466.7\,ms     & 522.267\,ms \\
    Adaptive Check   & 0.9\,\textmu s & 1.0\,\textmu s \\
    Total Prediction & 65.3\,ms      & 73.383\,ms \\
    VLM Call Ratio   & 14.0\,\%      & 14.0\,\% \\
    \end{tblr}
  }
\end{table}

\section{Conclusion}

This work presented a VLM based framework for GNSS spoofing detection in autonomous vehicles that audits behavior by comparing maneuvers inferred from synchronized camera and in vehicle sensors with maneuvers implied by GNSS. VLM model was fine-tuned using three stage pipeline to ensure visual cues and sensor traces alignment within a shared semantic space for robust maneuver prediction. This model was then validated using an independent field collected data in Tuscaloosa, USA. From this dataset we generated wrong turn, overshoot, and stop attack scenarios and used them to evaluate spoofing detection. We used adaptive inference policy to limit continuous model calls of the VLM to reduce the computational resources. Our VLM-based framework provides robust defense against adversarial attacks. Even if an adversary simultaneously compromises both GNSS signals and camera inputs using state-of-the-art spoofing techniques, the attack would fail. This resilience stems from VLMs' demonstrated ability to withstand various adversarial perception attacks, including shadow-based traffic sign perturbations, Dirty Road Patch lane attacks, and vehicle camouflage~\cite{mohajeransari2025natural}.

A major limitation of the current implementation is its computational and memory demand. The experiments were conducted on a development/evaluation server with a 48-core CPU and an NVIDIA A100 GPU, not an automotive-grade embedded compute platform; the peak GPU memory allocation was approximately 14 GB for LLaVA-Next-video-7B and 16 GB for NVILA-8B-Video, and individual VLM inference calls still introduce nontrivial latency. The adaptive inference policy reduces how often the VLM is invoked, but it does not reduce the memory or latency cost of each individual call, so it does not by itself close the gap to automotive-grade hardware. Closing this gap would require deployment-oriented compression, such as quantization, pruning, or distillation into a smaller task-specific model, including lightweight VLM backbones like TinyLLaVA~\cite{zhou2402tinyllava} or TinyGPT-V~\cite{yuan2312tinygpt}, combined with inference acceleration via TensorRT or ONNX Runtime and validation on automotive edge-computing hardware. An alternative path is an edge-cloud architecture, where VLM inference is offloaded to a remote server while the vehicle retains only the lightweight sensor-gating logic on board; this shifts the constraint from on-board compute to network latency and reliability, introducing its own deployment trade-offs.

Another limitation is that localization -drift attacks may gradually bias the GNSS position while preserving the same high-level semantic behavior. In such cases, only the vehicle’s position estimate (latitude, longitude) shifts slightly, but remains semantically consistent across GNSS, camera, and vehicle-side sensors; thereby, speed and yaw-rate variation may remain below the gating thresholds. A stronger mitigation is to combine the proposed behavior-level VLM detector with complementary metrics such as camera-based lane detection, lane-level map matching. Also, our approach assumes reliable operation of all sensor modalities without failures or synchronization issues. This creates vulnerability to sensor malfunctions or timing misalignment that could compromise detection accuracy. Future work should validate our approach against actual 
RF-level spoofing attacks using hardware-based signal generators to assess detection performance 
under authentic attack conditions and evaluate robustness against adversarially-optimized 
spoofing strategies. 

\printbibliography[title=References]

\end{document}